\documentclass[10pt,twocolumn,letterpaper]{article}

\usepackage[final]{cvpr}              
\usepackage{multirow}
\usepackage{bbding}
\usepackage{caption}
\usepackage[dvipsnames]{xcolor}

\definecolor{cvprblue}{rgb}{0.21,0.49,0.74}
\usepackage[pagebackref,breaklinks,colorlinks,citecolor=cvprblue]{hyperref}

\def\paperID{10220} 
\def\confName{CVPR}
\def\confYear{2024}
\def\datasetname{WildRGB-D}
\title{RGBD Objects in the Wild: \\ Scaling Real-World 3D Object Learning from RGB-D Videos}

\author{Hongchi Xia$^{1}$\footnotemark[1]
\quad
Yang Fu$^{2}$\footnotemark[1]
\quad 
Sifei Liu$^{3}$
\quad 
Xiaolong Wang$^{2}$\\
$^{1}$University of Illinois Urbana-Champaign
\quad
$^{2}$UC San Diego
\quad 
$^{3}$NVIDIA
}
\newcommand{\sumbullet}[1]{\textcolor{black}{#1}}
\newcommand{\customfootnotetext}[2]{{
\renewcommand{\thefootnote}{#1}
\footnotetext[0]{#2}}}
\begin{document}
\twocolumn[{%
\renewcommand\twocolumn[1][]{#1}%
\maketitle
\vspace{-10mm}
\captionsetup{type=figure}
\begin{center}
\includegraphics[width=0.8\textwidth]{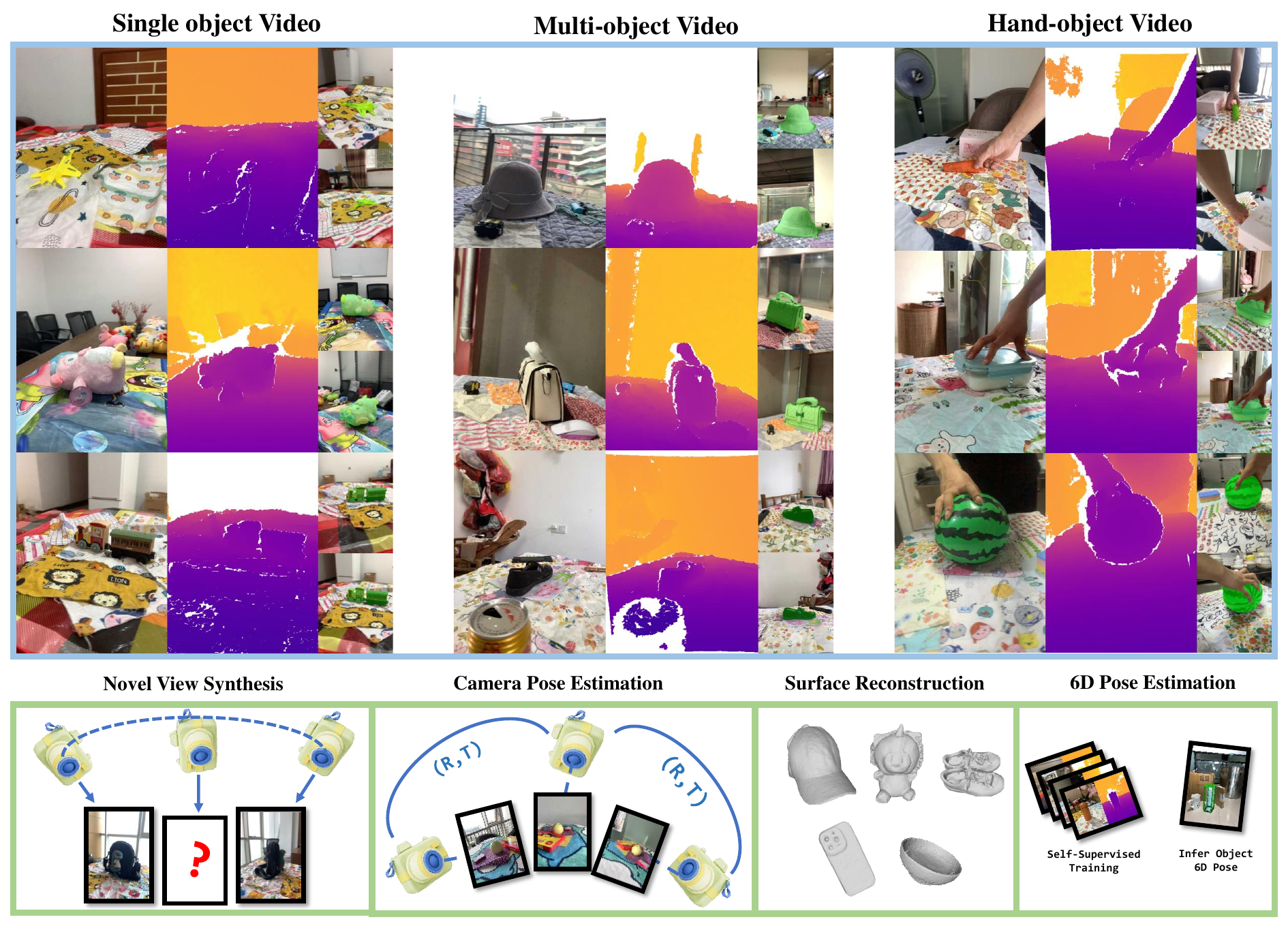}
\end{center}
\vspace{-5mm}
\captionof{figure}{\textbf{\datasetname~Dataset} contains almost 8500 recorded objects and nearly 20000 RGBD videos in 46 common categories with corresponding object masks and 3D point clouds.}
\label{fig:teaser}
}]
\customfootnotetext{*}{Equal contribution.}
\begin{abstract}
We introduce a new RGB-D object dataset captured in the wild called WildRGB-D. Unlike most existing real-world object-centric datasets which only come with RGB capturing, the direct capture of the depth channel allows better 3D annotations and broader downstream applications. WildRGB-D comprises large-scale category-level RGB-D object videos, which are taken using an iPhone to go around the objects in 360 degrees. It contains around 8500 recorded objects and nearly 20000 RGB-D videos across 46 common object categories. These videos are taken with diverse cluttered backgrounds with three setups to cover as many real-world scenarios as possible: (i) a single object in one video; (ii) multiple objects in one video; and (iii) an object with a static hand in one video. The dataset is annotated with object masks, real-world scale camera poses, and reconstructed aggregated point clouds from RGBD videos. We benchmark four tasks with WildRGB-D including novel view synthesis, camera pose estimation, object 6d pose estimation, and object surface reconstruction. Our experiments show that the large-scale capture of RGB-D objects provides a large potential to advance 3D object learning. Our project page is \url{https://wildrgbd.github.io/}.


\end{abstract}   
\begin{table*}
   \centering
   \begin{tabular}{c}
   \includegraphics[width=0.8\textwidth]{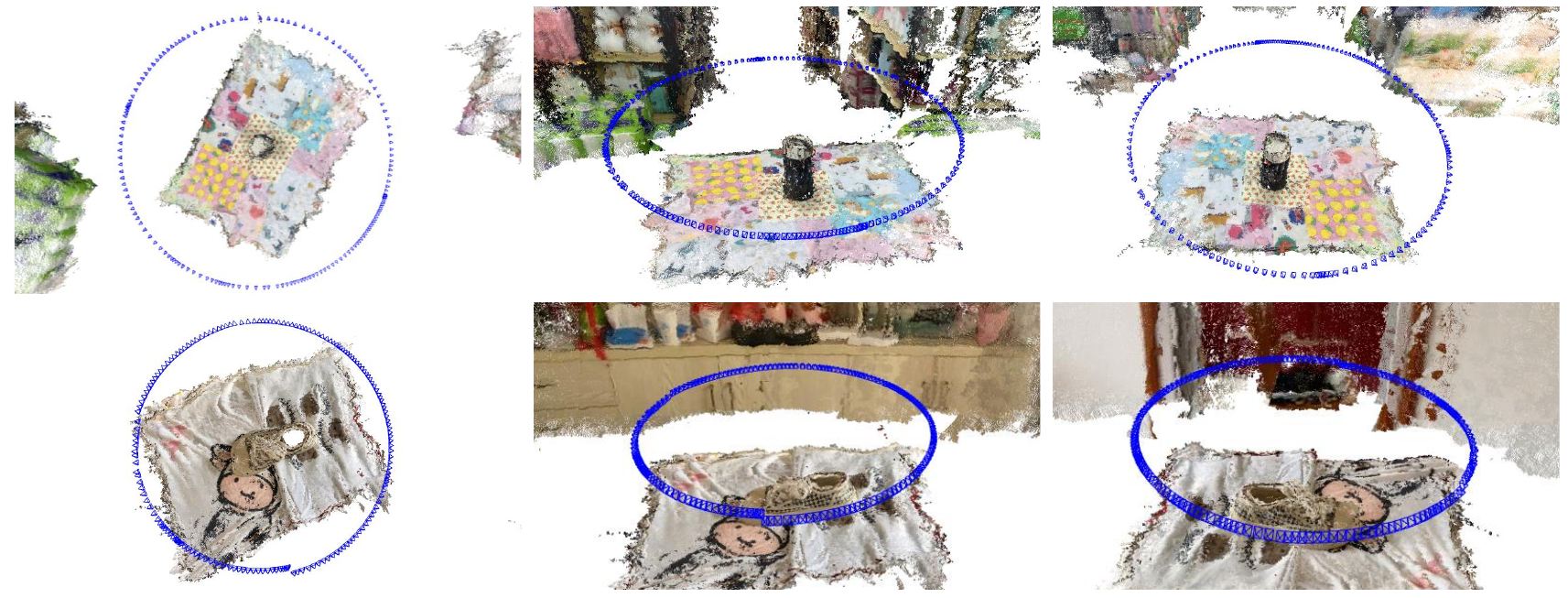}
   \end{tabular}
   
   \vspace{-5px}
   \captionof{figure}{{\bf The camera poses trajectory in \datasetname~Dataset.} We visualize the corresponding camera in each scene of our dataset, showing that our dataset is featured in 360 degree full and dense multi-view camera poses.}
    \label{fig:traj}
\end{table*}

\section{Introduction}
\label{sec:intro}



The recent advancement of computer vision has been largely relying on the scaling of training data~\cite{he2015deep,radford2021learning}. The same success in data-driven approaches has been recently adopted to 3D object modeling with new large 3D object-centric dataset collection~\cite{liu2023zero1to3,qi2017pointnet,gao2022get3d}. Most of the large datasets are synthetic 3D data~\cite{chang2015shapenet,fu20213d,wu20153d,collins2022abo} and a mix of synthetic data and real-world object scans~\cite{deitke2022objaverse}, given it is much less labor intensive for scaling by rendering from simulation. However, it remains a big challenge to apply the model trained in simulation data to the real world. This is not only because the synthetic data has less realistic texture and shape, but also due to it is very hard to model the cluttered background and the natural light comes with it in simulation.

To make deep learning with 3D objects applicable in the real world, researchers have made efforts to collect real-world multiview object data~\cite{reizenstein2021common,ahmadyan2021objectron}. For example, the CO3D dataset~\cite{reizenstein2021common} contains 19K object videos across 50 categories. However, due to the lack of depth, they require the use of COLMAP~\cite{schoenberger2016sfm} to provide 3D annotations, which only works for $20\%$ of the collected videos. Collecting the depth channel part of the data is not only useful for more accurate 3D ground-truth annotations, but also provides very useful information for downstream applications such as object 6D pose estimation and novel view synthesis. The OmniObject3D dataset~\cite{wu2023omniobject3d} provides both object videos and a separate scanning of the objects. However, the collected videos do not come with the depth channel inputs and they are mostly taken with clean backgrounds. The Wild6D dataset~\cite{fu2022categorylevel} is one of the few recent efforts to collect RGB-D object videos taken in the wild. However, it only contains 6 categories of data and covers relatively smaller ranges of object views. 

In this paper, we propose to collect a new dataset that contains large-scale RGB-D object videos across diverse object categories and presented in the wild. Our dataset, namely \datasetname, covers 8500 tabletop objects across 44 categories in 20K videos. The videos are taken using iPhones to go around the objects in 360 degrees (see Figure~\ref{fig:traj} for visualization). Examples of the dataset are shown in Figure~\ref{fig:teaser}. There are three types of videos: (i) \emph{Single object video} where there is only one object presented on the table; (ii) \emph{Multi-object video} where there are multiple objects presented at the same time; and (iii) \emph{Hand-object video} where there is a static human hand grasping the object. More video types add variety, creating occlusion for objects in scenes, which are worthy study cases in some tasks. The collection of the \datasetname~ dataset not only considers the cluttered background in the real world, but also the common scenarios where the objects are occluded by human hands.

We perform automatic annotations for \datasetname. With RGB-D capturing, we can apply the Simultaneous Localization and Mapping~\cite{durrant2006simultaneous,schops2019bad} (SLAM) algorithms, and exploit the RGB images and depth information from the depth sensor of mobile phones to reconstruct the \emph{3D camera poses} in real-world scale and aggregated \emph{3D point clouds}. Additionally, center \emph{object segmentation masks} can be attained by bounding-box detection using text prompt of object category in Grounding-DINO~\cite{liu2023grounding}, segmentation using Segment-Anything~\cite{kirillov2023segment} and mask tracking using XMem~\cite{cheng2022xmem}, which are largely integrated into \cite{cheng2023segment,yang2023track}.

To exploit the potential of our dataset, we benchmark it in four downstream tracks: 

\textbf{(i) Novel view synthesis.} We evaluate various algorithms based on NeRF~\cite{mildenhall2020nerf} which is optimized in a single scene, and generalizable NeRF which is trained in a category-level. With the help of depth information when training NeRFs, we can achieve consistently improved results. This offers a new platform for evaluating view synthesis approaches using RGB or RGB-D data. 

\textbf{(ii) Camera pose estimation.} 
We adopt different pose estimation approaches~\cite{zhang2022relpose,lin2023relpose++} to evaluate their capability of estimating relative camera poses in a sparse setting. We validate their generalizable ability through training on a partial of all categories and testing on unseen ones. We observe remarkable generalization performance on unseen categories, which indicates our large-scale category-level dataset can serve as a training source for generalizable camera pose estimation.

\textbf{(iii) Object surface reconstruction.}
We conduct object surface reconstruction in our dataset with RGB or RGB-D videos and object masks through Instant-NGP~\cite{muller2022instant} and Neusfacto~\cite{Yu2022SDFStudio}. Results show that our depth information endow reconstruction with more accurate precision and SDF-based algorithm~\cite{Yu2022SDFStudio} performs better in this setting.

\textbf{(iv) Object 6D pose estimation.} 
We exploit the self-supervised algorithm in category-level object 6D pose estimation~\cite{zhang2022self} with large-scale RGB-D images in our dataset and then evaluate the pre-trained model on the Wild6D~\cite{fu2022categorylevel} test set. We show our dataset can facilitate 6D pose estimation even without training labels, and we also study its generalization ability to the out-of-distribution test set.

\section{Related Work}
\label{sec:related}
\begin{table}[t]
    \resizebox{0.48\textwidth}{!}{
    \centering
    \begin{tabular}{l|ccccccc}
    \toprule
        Dataset & Real & Multi-View & Depth Src. & 3D GT & Video & \# Cats & \# Objs \\
        \midrule
        ShapeNet~\cite{chang2015shapenet} &  & none & CAD & mesh & none & 55 & 51k \\
        ModelNet~\cite{wu20153d} &  & none & CAD & mesh & none & 40 & 12k \\
        3D-Future~\cite{fu20213d} &  & none & CAD & mesh & none & 34 & 16k \\
        ABO~\cite{collins2022abo} &  & none & CAD & mesh & none & 63 & 8k \\
        DTU~\cite{aanaes2016large} & \Checkmark  & limited & COLMAP & mesh & RGB & N/A & 124 \\
        CO3D~\cite{reizenstein2021common} & \Checkmark & full & COLMAP & pcl & RGB & 50 & 19k \\
        {MVImgNet~\cite{yu2023mvimgnet}} & \Checkmark & {limited} & {COLMAP} & {pcl} & {RGB} & {238} & {220k} \\
        Objectron~\cite{ahmadyan2021objectron} & \Checkmark & limited & COLMAP & pcl & RGB & 9 & 15k \\
        GSO~\cite{downs2022google}  & \Checkmark & none & scanner & mesh & none & 17 & 1k \\
        OmniObject3D~\cite{wu2023omniobject3d} & \Checkmark & full & scanner &  mesh & RGB & 190 & 6k \\
        Choi et al.~\cite{choi2016large} & \Checkmark & limited & sensor & mesh* & RGBD & 9 & 2k \\
        Wild6D~\cite{fu2022categorylevel} & \Checkmark & limited & sensor & pcl & RGBD & 5 & 1.8k \\
        \textbf{Ours} & \Checkmark & full & sensor & pcl & RGBD & 44 & 8.5k \\
    \bottomrule
    \end{tabular}
    }
    \caption{\textbf{Comparison of \datasetname~dataset with other 3D object dataset.} {Some datasets don't provide video and we mark in ``none''. Some only cover partial angles, which is marked in ``limited''. Asterisk(*) means partial annotations. Depth Src. means where the depth information comes from, including CAD models, COLMAP, scanner devices and depth sensor in iPhones. pcl is the abbreviation of point cloud.\\
    }}
    \label{tab:dscmp}

\end{table}
\begin{table*}
   \centering
   \resizebox{\textwidth}{!}{
   \begin{tabular}{c}
   \includegraphics[width=\textwidth]{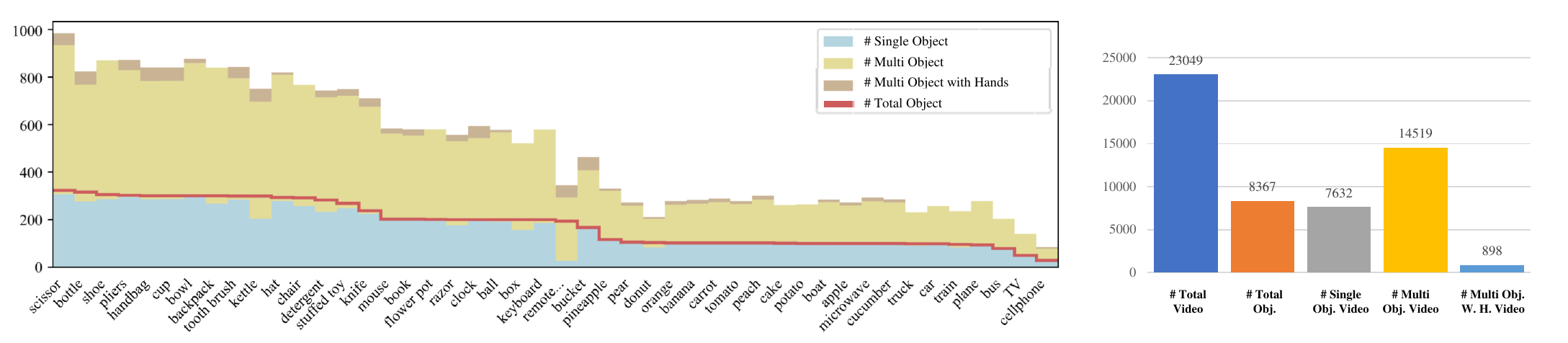}
   \end{tabular}
   
   }
   \vspace{-10px}
   \captionof{figure}{{\bf Statistics of \datasetname~Dataset} {list the total and per-category number of objects and different types of videos.}}
    \label{fig:statistic}
\end{table*}
\begin{table}
   \centering
   \resizebox{\linewidth}{!}{
   \begin{tabular}{c}
   \includegraphics[width=\linewidth]{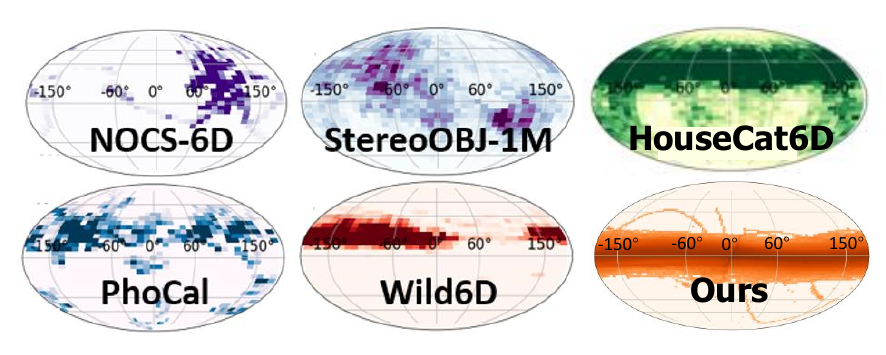}
   \end{tabular}
   
   }
   \vspace{-5px}
   \captionof{figure}{{\bf Distribution visualization} {of different kinds of Object 6D pose dataset and \datasetname~ dataset. We observe obvious disparity between Wild6D and Our dataset. Visualizations of ~\cite{wang2019normalized, wang2022phocal, liu2022stereobj1m, fu2022categorylevel, jung2023housecat6d} are from ~\cite{jung2023housecat6d}.}}
    \label{fig:distvis}
\end{table}
\paragraph{3D Object Datasets}
One representative kind of 3D object dataset is the 3D synthetic dataset, like  ShapeNet~\cite{chang2015shapenet} and ModelNet40~\cite{wu20153d}, which consist of category-level objects. 3D-FUTURE~\cite{fu20213d} and ABO~\cite{collins2022abo} datasets are typical of higher quality mesh with textures. \cite{lim2013parsing} and \cite{tulsiani2016learning} introduce real-world category-specific object datasets that mainly focus on birds and chairs respectively. DTU~\cite{aanaes2016large} and BlendedMVS~\cite{yao2020blendedmvs} are datasets designed for multi-view stereo and lack category-level classification. Objectron~\cite{ahmadyan2021objectron} provides rich annotations but only partial videos are fully 360 degree covered. CO3D~\cite{reizenstein2021common} is a large-scale category-level dataset that annotates camera poses and depths with COLMAP~\cite{schonberger2016structure}, which doesn't provide depths in real-world scale, and MVImgNet~\cite{yu2023mvimgnet} is also a dataset similar to CO3D. Pascal3D~\cite{xiang2014beyond} contains real-world 3D objects with pose annotations and CAD models in limited categories. Datasets collected with specialized hardware (scanner, dome, etc.)  like GSO~\cite{downs2022google} and OmniObject3D~\cite{wu2023omniobject3d} have more accurate 3D geometry models and rendered depths from them. However, they don't have RGBD wild object videos collected and lack real captured depths as well as background depths. In the aspect of RGBD object datasets, Wild6D~\cite{fu2022categorylevel} features RGBD image sequences and 6D pose annotations while lacking full 360 coverage and category types. Choi et al.~\cite{choi2016large} proposes RGBD object-centric datasets in 44 categories, but with limited camera annotations. As a comparison, our proposed \datasetname~ dataset contains almost 8500 recorded objects and nearly 20000 RGBD videos recorded all 360 degrees in 46 common categories from well-known 2D datasets, all with real-world scale camera poses and object mask annotations as well as aggregated point clouds. We present the detailed comparison in Tab.~\ref{tab:dscmp}.





\paragraph{Neural Radiance Field and Object Surface Reconstruction}
Neural Radiance Field (NeRF)~\cite{mildenhall2020nerf} is a kind of scene representation based on MLPs. It takes in sampled points along each ray and outputs the density and color of each point, which are then aggregated by volume rendering to synthesize views. \cite{barron2021mip,barron2022mip,mildenhall2022nerf,verbin2022ref} propose new changes to the original NeRF to improve the visual quality and \cite{chen2022tensorf,fridovich2022plenoxels,muller2022instant,sun2022direct,Lin_2022_CVPR} advance the NeRF efficiency. In order to generalize the NeRF representations to other scenes, ~\cite{yu2021pixelnerf,chen2021mvsnerf,wang2021ibrnet,henzler2021unsupervised,liu2022neural} learn latent 3D representations and priors from a bunch of existed scenes to help synthesize views across different scenes. Derived from original NeRF, \cite{wang2021neus, yariv2021volume, Yu2022SDFStudio, darmon2022improving, oechsle2021unisurf, wu2022voxurf, wang2023neus2} leverage Sign Distance Function (SDF) and represent the 3D scene by implicit surface, which has a more clear object boundary definition. Recently, 3D Gaussian Splating~\cite{kerbl20233d} has become a competitive alternative to NeRF. \datasetname~ dataset comprises various category-level objects and scenes on a large scale, which is suitable for novel view synthesis benchmarks and helps boost more mature reconstruction algorithms and generalizable 3D scene representations.

\paragraph{Camera Pose Estimation}
Given dense image views, mature algorithms of SfM~\cite{schonberger2016structure} and SLAM~\cite{durrant2006simultaneous} can estimate camera poses well by computing visual matches~\cite{lucas1981iterative}, verifying through RANSAC~\cite{fischler1981random} and optimizing via bundle adjustment~\cite{triggs2000bundle}. However, in a sparse camera view setting, camera pose estimation remains a challenging task. Some approaches~\cite{teed2021droid,wang2017deepvo} leverage RNN or adopt auto-regression~\cite{yang2020d3vo} targeting at SLAM applications. For category-agnostic sparse view camera pose estimation, ~\cite{melekhov2017relative,rockwell20228} adopt a direct regression approach. ~\cite{jiang2022few} estimates 6D pose upon training on synthetic dataset. Energy-based method~\cite{zhang2022relpose} estimates distributions over relative rotations and ~\cite{lin2023relpose++} incorporates multi-view context to estimate camera 6D pose. Bundle adjustment gets learned after predictions in ~\cite{sinha2023sparsepose} to refine the estimated poses. In \datasetname~ dataset, with full 360-degree multi-view videos, the sparse view camera pose estimation setting is easily accessible, enabling our dataset to serve as a large-scale training database for these algorithms.

\paragraph{Object 6D Pose Estimation}
In the setting of category-level 6D pose estimation, algorithms predict object poses in the same category and meet with various intra-class shapes. ~\cite{wang2019normalized} predicts 6D pose using Umeyama algorithm~\cite{umeyama1991least} with NOCS map estimation and ~\cite{tian2020shape,chen2020learning,fu2022categorylevel} follow up to learn more accurate NOCS representations. Other algorithms learn to estimate 6D pose through direct regression~\cite{chen2021fs,lin2021dualposenet}, keypoint location estimations~\cite{lin2022single} and so on. Apart from supervised learning, self-supervision emerges due to the high cost of annotations. One approach~\cite{chen2020category,he2022towards,gao20206d,you2022cppf} is to adapt sim-to-real upon the pre-trained model on synthetic data. Another one~\cite{fu2022categorylevel,manhardt2020cps++,peng2022self} resorts to semi-supervised training. \cite{zhang2022self} proposes cycles across 2D-3D space learned correspondence, which enables training using only in-the-wild RGBD images without any annotations and is compatible with our dataset. With large-scale category-level RGBD wild object images for self-supervised learning, our dataset has the potential to boost future developments in this field.
\section{The \datasetname ~Dataset}
\begin{table*}
   \centering
   \resizebox{\textwidth}{!}{
   \begin{tabular}{c}
   \includegraphics[width=\textwidth]{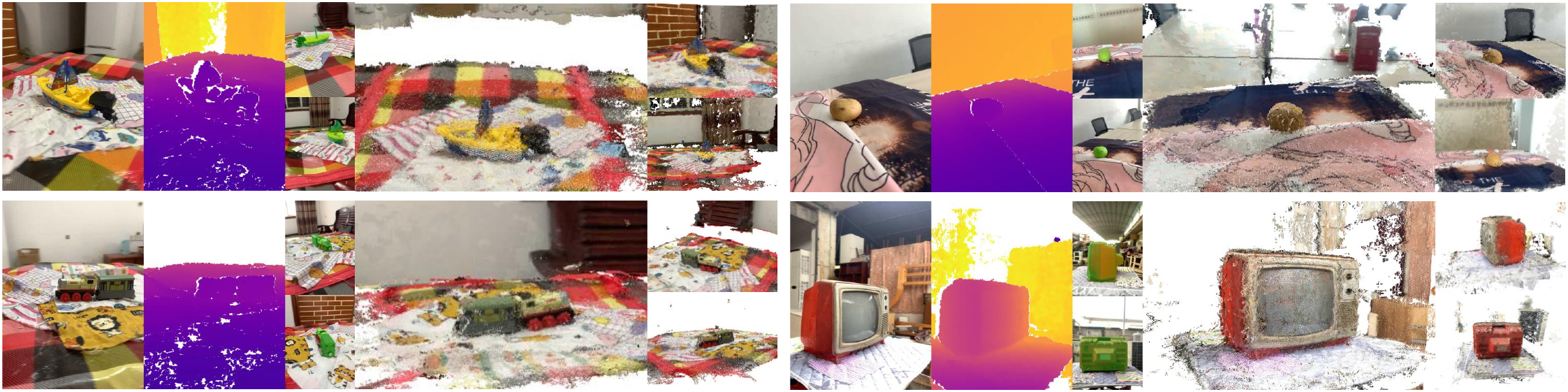}
   \end{tabular}
   
   }
   \vspace{-5px}
   \captionof{figure}{{\bf Point cloud reconstruction of objects in \datasetname~Dataset.} We reconstruct the aggregated point cloud of the scene by leveraging existed 3D annotations of camera poses and depth images.}
    \label{fig:pcd}
\end{table*}

\subsection{Data Collections, Processing, and Annotation}
\paragraph{Data Collections} In order to collect RGBD video on a large scale expediently and economically, we record with the help of an iPhone front camera using Record3D~\footnote{https://record3d.app/} App and rotate the camera around the object so that full 360-degree views of objects are captured with RGB images and the corresponding depth images. Camera rotating speed is controlled equably by our collection setup to ensure less blur in videos. We select 46 common categories from well-known 2D datasets~\cite{deng2009imagenet,lin2014microsoft}. We record three videos for every selected object, which are composed of single-object video, multi-object video, and hand-object video. Every recorded video has been checked and some are left behind due to poor SLAM camera pose estimation. Details of \datasetname~Dataset are listed in Fig.~\ref{fig:statistic}.
\vspace{-12px}
\paragraph{Generating Camera Poses and 3D Point Cloud}
Our \datasetname~dataset has 3D annotations including camera poses in real-world scale, scene point clouds, and central object masks. 
In order to attain real-world scale camera poses, 
instead of relying on COLMAP~\cite{schonberger2016structure} to first generate camera poses and then the depths using the poses, we generate more accurate camera poses with the mature RGBD Simultaneous Localization and Mapping~\cite{durrant2006simultaneous,schops2019bad} (SLAM) algorithm, which leverages our captured depths. Additionally, it has the capability of exploiting the RGB images and depth information from the depth sensor of mobile phones to reconstruct the 3D camera poses in real-world scale, which is different from COLMAP depths, which are not in real-world scale.
It enables us to simply project the depth images and gain aggregated 3D point clouds (see Figure~\ref{fig:pcd}). 
Then we manually check the quality of the aggregated 3D point cloud and exclude videos in which SLAM fails to get accurate camera poses. To increase the probability of getting correct SLAM results for each video, we adopt two kinds of SLAM algorithms including BAD SLAM~\cite{schops2019bad} and SLAM implementation from Open3D~\cite{Zhou2018}, which increase our successful rate to over 90\%.
\vspace{-12px}

\paragraph{Generating Central Object Masks} We perform central object mask segmentation through a series of methods. Instead of the classic PointRend~\cite{kirillov2020pointrend} algorithm, we leverage the novel segmentation tool Segment-Anything (SAM)~\cite{kirillov2023segment}. We attain the prompt for SAM using Grounding-DINO~\cite{liu2023grounding}, which generates a bounding box for SAM according to the category text prompt. After attaining the mask segmentation of the first frame in the video, XMem~\cite{cheng2022xmem} is applied to track the mask in the video. The masking pipeline is largely integrated into \cite{cheng2023segment,yang2023track}.
\vspace{-4px}

\subsection{Statistics and Distribution}
In \datasetname~dataset collections, we recorded 8500 objects and 3 videos for each one. After excluding those SLAM-failed videos, we have 8367 objects in 23049 videos in our dataset (maintaining rates are 99.3\%/91.0\%). The selected videos contain 33.1\% single object videos, 63.0\% multi-object videos, and 3.9\% hand-object videos. Details of \datasetname~ dataset are listed in Fig.~\ref{fig:statistic}.
\section{Experiments}
\label{sec:exp}
\begin{table}[t]
    \resizebox{0.48\textwidth}{!}{
    \centering
    \begin{tabular}{l|cccc}
    \toprule
        Method &  PSNR$\uparrow$/SD & SSIM$\uparrow$/SD & LPIPS$\downarrow$/SD & MAE$\downarrow$/SD \\
        \midrule
         NeRF~\cite{mildenhall2020nerf}  & 23.03/1.50  &	0.690/0.072 	& 0.390/0.075  & 0.306/0.109  \\
         NeRF (w mask)  & 34.65/4.44  &	0.943/0.077  	& 0.031/0.032   & 0.029/0.019  \\
         Mip-NeRF 360~\cite{barron2022mip}  & \textbf{23.84}/1.60  &	\textbf{0.762}/0.063  &	0.280/0.067  & \textbf{0.185}/0.068  \\
         Mip-NeRF 360 (w mask)  &  35.60/4.51  &	\textbf{0.949}/0.077  &	0.024/0.025    & \textbf{0.020}/0.015 \\
         Instant-NGP~\cite{muller2022instant}  &  23.67/2.07  &	0.745/0.063  &	\textbf{0.257}/0.070   & 0.366/0.105 \\
         Instant-NGP (w mask)  &  \textbf{35.65}/5.20  	& 0.946/0.077  &	\textbf{0.021}/0.031   & 0.068/0.074  \\
    \bottomrule
    \end{tabular}
    }
    \caption{\textbf{Single-scene NVS results.} {Average of four metrics w and w/o masks across all training dataset are reported with their standard deviation (SD).}}
    \label{tab:single_nvs}
\end{table}
\begin{table}[t]

    \resizebox{0.48\textwidth}{!}{
    \centering
    \begin{tabular}{l|c|cccc}
    \toprule
        Method & Level & PSNR$\uparrow$/SD & SSIM$\uparrow$/SD & LPIPS$\downarrow$/SD & MAE$\downarrow$/SD\\
        \midrule
         Pixel-NeRF~\cite{yu2021pixelnerf}  & \multirow{3}{*}{Easy} & 20.28/0.65    &  0.645/0.043   &  0.495/0.074   & \textbf{0.355}/0.120  \\
         MVSNeRF~\cite{chen2021mvsnerf}  &  & 19.95/1.00   & 0.663/0.036  & \textbf{0.351}/0.066   & 0.370/0.100 \\
         IBRNet~\cite{wang2021ibrnet}  &  & \textbf{20.93}/0.98   & \textbf{0.711}/0.031  & 0.395/0.153   & - \\
        \midrule
        Pixel-NeRF~\cite{yu2021pixelnerf}  & \multirow{3}{*}{Middle} & 18.76/0.50  & 0.572/0.064  & 0.534/0.047   &  \textbf{0.299}/0.057 \\
         MVSNeRF~\cite{chen2021mvsnerf}  &  & 18.75/0.74   & 0.601/0.069  & 0.363/0.036   & 0.345/0.102 \\
         IBRNet~\cite{wang2021ibrnet}  &  & \textbf{19.77}/1.01   & \textbf{0.663}/0.071  & \textbf{0.362}/0.063  & - \\
        \midrule
        Pixel-NeRF~\cite{yu2021pixelnerf}  & \multirow{3}{*}{Hard} & 17.23/0.66  &  0.521/0.035  & 0.624/0.054 &  \textbf{0.383}/0.121 \\
         MVSNeRF~\cite{chen2021mvsnerf}  &  & 17.13/0.89   & 0.564/0.043   &  \textbf{0.425}/0.045  & 0.502/0.260 \\
         IBRNet~\cite{wang2021ibrnet}  & & \textbf{17.92}/1.12  & \textbf{0.614}/0.056  & 0.439/0.069   & - \\
    \bottomrule
    \end{tabular}
    }
    \caption{\textbf{Cross-scene NVS results.} {Average of four metrics across all categories in training dataset are reported. We report metrics of three difficulty level respectively. Entries marked in - are not provided.}}
    \label{tab:cross_nvs}

\end{table}

\begin{table}[t]
    \resizebox{0.48\textwidth}{!}{
    \centering
    \begin{tabular}{l|cccc}
    \toprule
        Method &  PSNR$\uparrow$/SD & SSIM$\uparrow$/SD & LPIPS$\downarrow$/SD & MAE$\downarrow$/SD \\
        \midrule
         Instant-NGP~\cite{muller2022instant}  & 23.67/2.07  & 0.745/0.063  & 0.257/0.070  & 0.366/0.105    \\
         Instant-NGP (depth sup.)  & \textbf{24.60}/2.13 & \textbf{0.759}/0.062  & \textbf{0.239}/0.066  & \textbf{0.108}/0.057 \\
        \midrule
         Pixel-NeRF~\cite{yu2021pixelnerf}  & 18.53/1.21 & 0.568/0.067 & 0.556/0.073 &  0.336/0.099 \\
         Pixel-NeRF (depth sup.)  & \textbf{19.10}/1.21 & \textbf{0.605}/0.060 & \textbf{0.499}/0.064  & \textbf{0.147}/0.087 \\
         MVSNeRF~\cite{chen2021mvsnerf}  & 18.43/1.30 & 0.600/0.065 & 0.381/0.054 & 0.400/0.182 \\
         MVSNeRF (depth sup.)  & \textbf{18.44}/1.29 & 0.600/0.065  & 0.381/0.054 & \textbf{0.397}/0.186 \\
    \bottomrule
    \end{tabular}
    }
    \caption{\textbf{Depth Supervised NVS and depth estimation results.} {Average of four metrics w and w/o depth supervision across all training dataset are reported with their standard deviation (SD).}}
    \label{tab:depth_sup_nvs}

\end{table}
\begin{table}
   \centering
   \resizebox{\linewidth}{!}{
   
   \begin{tabular}{c|ccc}
   Target & 
   NeRF~\cite{mildenhall2020nerf} & 
   Mip-NeRF 360~\cite{barron2022mip}  & 
   Instant-NGP~\cite{muller2022instant} \\
   \includegraphics[width=0.4\linewidth]{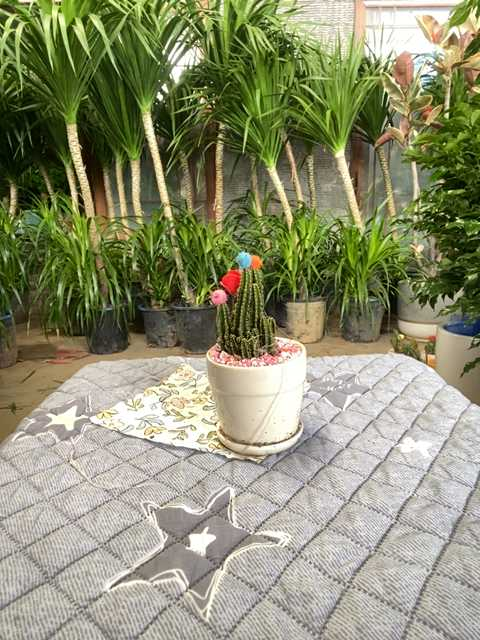} &
   \includegraphics[width=0.4\linewidth]{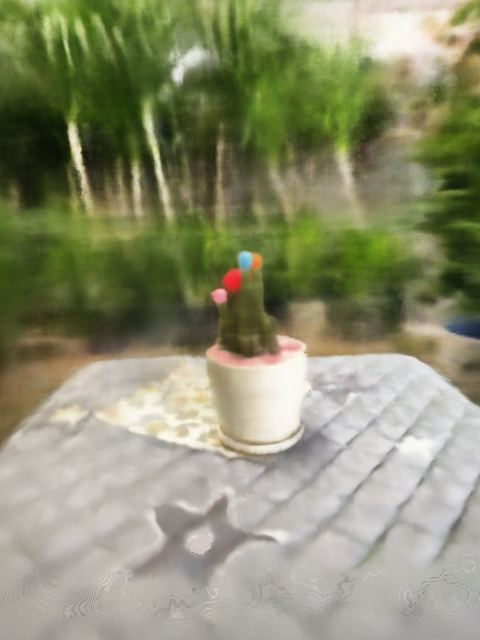} &
   \includegraphics[width=0.4\linewidth]{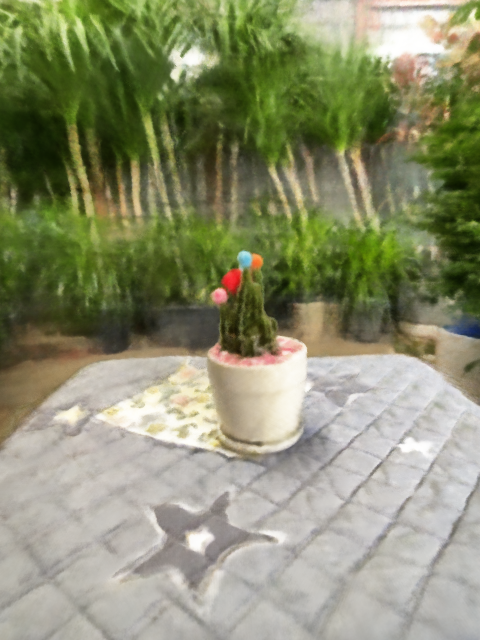} &
   \includegraphics[width=0.4\linewidth]{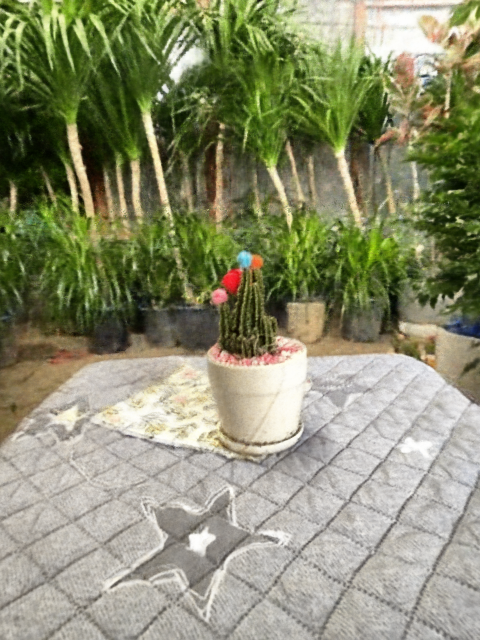} \\
      \includegraphics[width=0.4\linewidth]{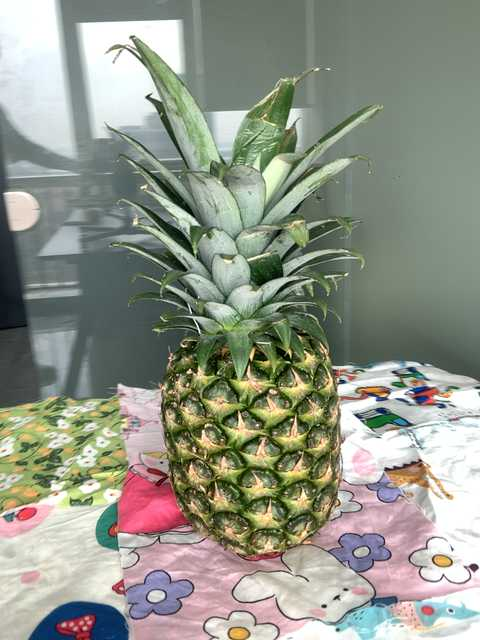} &
   \includegraphics[width=0.4\linewidth]{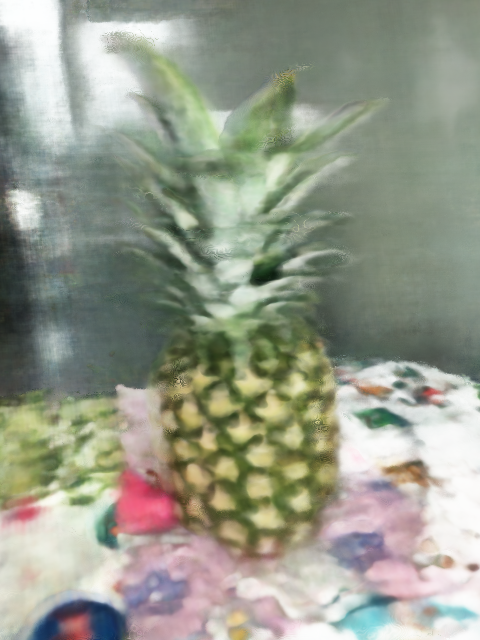} &
   \includegraphics[width=0.4\linewidth]{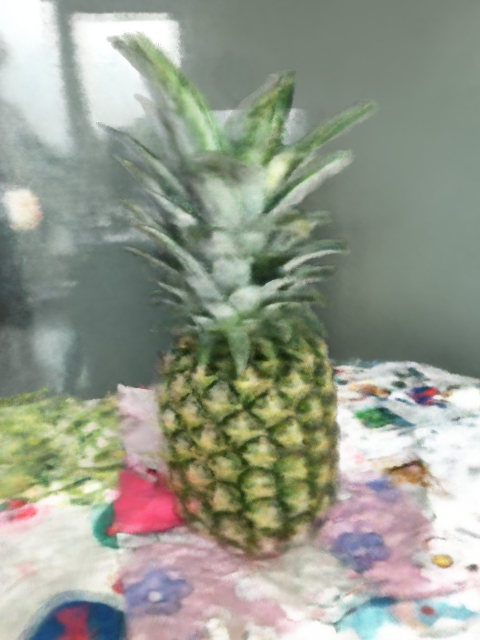} &
   \includegraphics[width=0.4\linewidth]{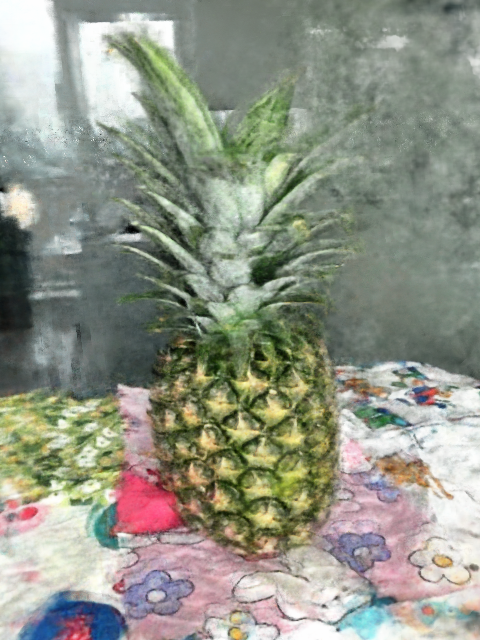} \\
      \includegraphics[width=0.4\linewidth]{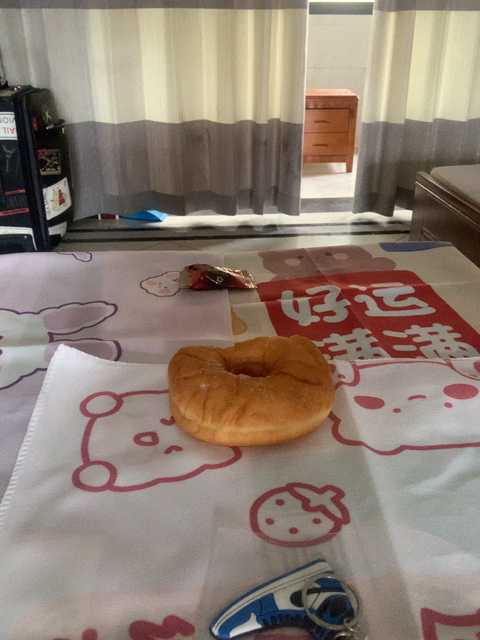} &
   \includegraphics[width=0.4\linewidth]{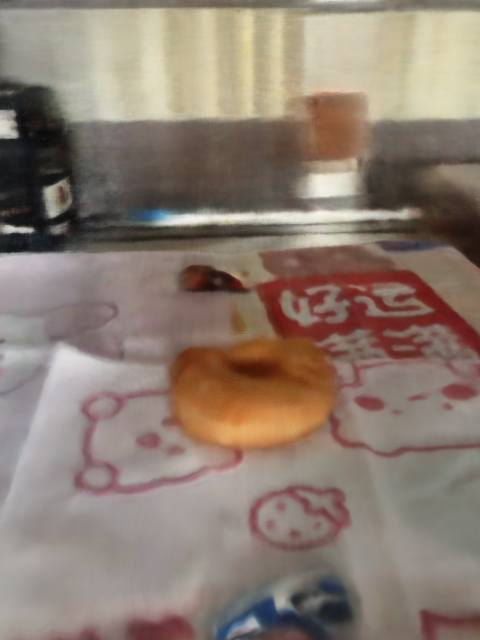} &
   \includegraphics[width=0.4\linewidth]{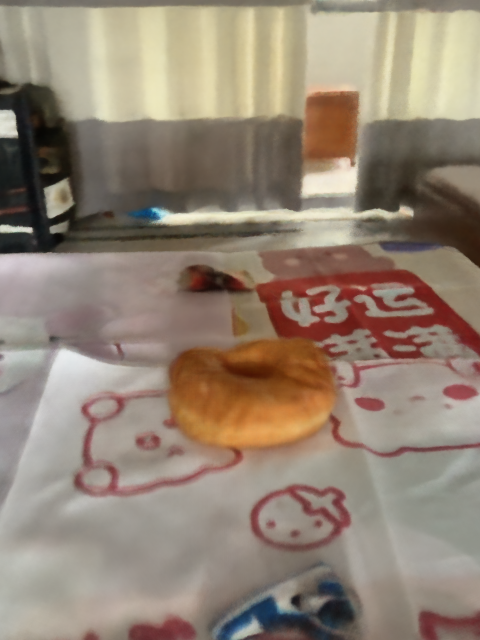} &
   \includegraphics[width=0.4\linewidth]{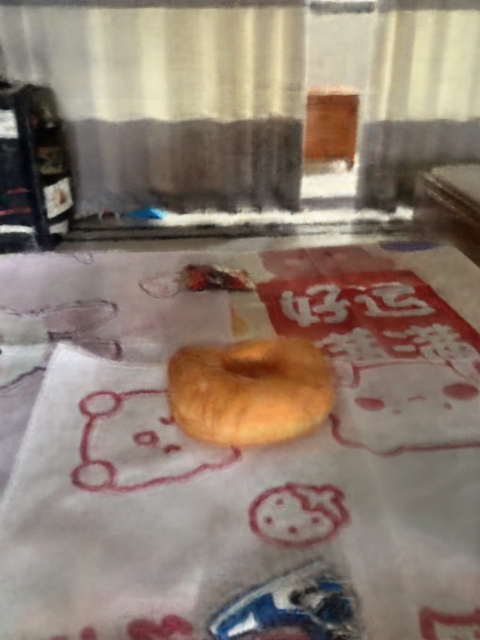}
   \end{tabular}
   }
   \vspace{-5px}
   \captionof{figure}{{\bf Novel view synthesis visualization} {of different kinds of NeRF methods: NeRF~\cite{mildenhall2020nerf}, Mip-NeRF 360~\cite{barron2022mip} and Instant-NGP~\cite{muller2022instant}.}}
    \label{fig:nvs}
\end{table}
\subsection{Novel View Synthesis}
In this section, we conduct multiple experiments towards methods concerning novel view synthesis (NVS) in the following three scenarios: 1) \textbf{Single-Scene NVS}, where we train NeRF-based methods \cite{mildenhall2020nerf, barron2022mip, muller2022instant} on a single scene with only RGB image sequence. 2) \textbf{Cross-Scene NVS}, where we learn category-level scene representations to generalize into other scenes with Generalizable NeRFs~\cite{yu2021pixelnerf, chen2021mvsnerf, wang2021ibrnet}. 3) \textbf{Depth Supervised NVS}, where we conduct NVS experiments with depth image priors in our dataset to study the potential that depth information will endow to NVS tasks.
\vspace{-12px}
\paragraph{Single-Scene NVS}
We select ten scenes from each category and uniformly sample images as validation split. We choose NeRF~\cite{mildenhall2020nerf}, Mip-NeRF 360~\cite{barron2022mip} and Instant-NGP~\cite{muller2022instant} for evaluations. Results are shown in Tab.~\ref{tab:single_nvs}. We report the average PSNR, SSIM~\cite{wang2004image}, LPIPS~\cite{zhang2018unreasonable} and rendering depths Mean Average Error (MAE) compared with our sensor-collected depths across all categories. We also report metrics only related to the NVS quality of central objects using object masks. Results show that Mip-NeRF 360 and Instant-NGP outperform original NeRF in terms of visual quality metrics. NeRF-based methods perform better when we only concern with the recovery of central objects under object masks. What's more, Mip-NeRF 360 performs best in learning single-scene geometry. Visualization can be found in Figure~\ref{fig:nvs}. \sumbullet{In brief, our dataset offers extensive categories and scenes for in-depth NVS experiments.}
\vspace{-12px}
\paragraph{Cross-Scene NVS}
Apart from single-scene optimizations, we also evaluate Generalizable NeRFs: Pixel-NeRF~\cite{yu2021pixelnerf}, MVSNeRF~\cite{chen2021mvsnerf} and IBRNet~\cite{wang2021ibrnet} in the cross-scene setting. For each category in our dataset, we select the same test scenes as single-scene NVS experiments and train in the remaining scenes of the same category to learn per-category latent representations. We divide the 46 categories into three difficulty levels and report the average metrics of each level. For evaluation, we use three source views to synthesize novel views. We report PSNR, SSIM and LPIPS to measure visual quality and depth MAE to measure the learned geometry quality.
From Tab.~\ref{tab:cross_nvs}, we observe that IBRNet outperforms in all three difficulty levels in terms of visual quality. Additionally, learned geometry quality isn't highly correlated with the rendering visual quality in novel views. \sumbullet{To sum up, our dataset provides great potential in learning category-level cross-scene NVS methods.}
\begin{table}
   \centering
   \resizebox{\linewidth}{!}{
   
   \begin{tabular}{c}
   \includegraphics[width=\linewidth]{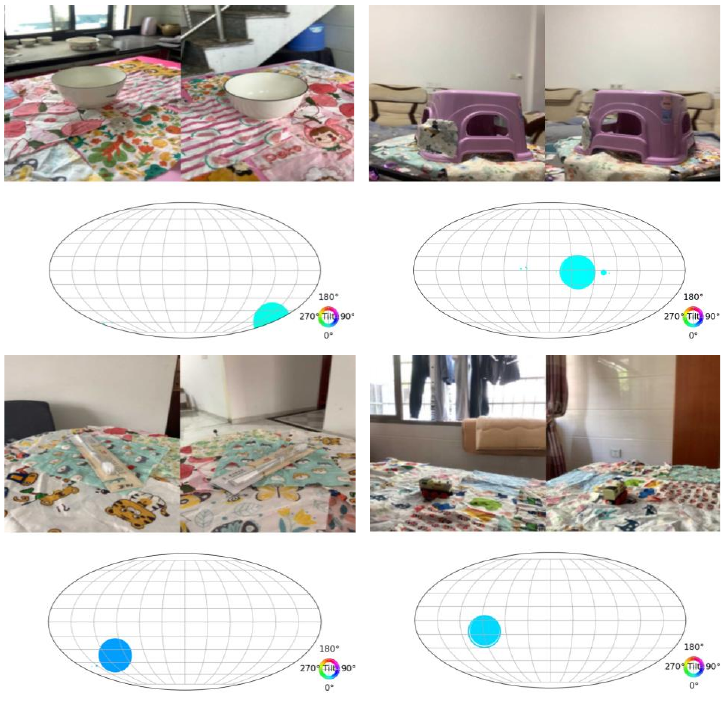} 
   \end{tabular}
   }
   \vspace{-5px}
   \captionof{figure}{{\bf Relpose++~\cite{lin2023relpose++} pair-wise evaluation visualization.} {We show every image pair with its  relative rotation predicted by Relpose.}}
    \label{fig:relpose}
\end{table}
\begin{table*}
   \centering
   \resizebox{\textwidth}{!}{
   \begin{tabular}{c}
   \includegraphics[width=\textwidth]{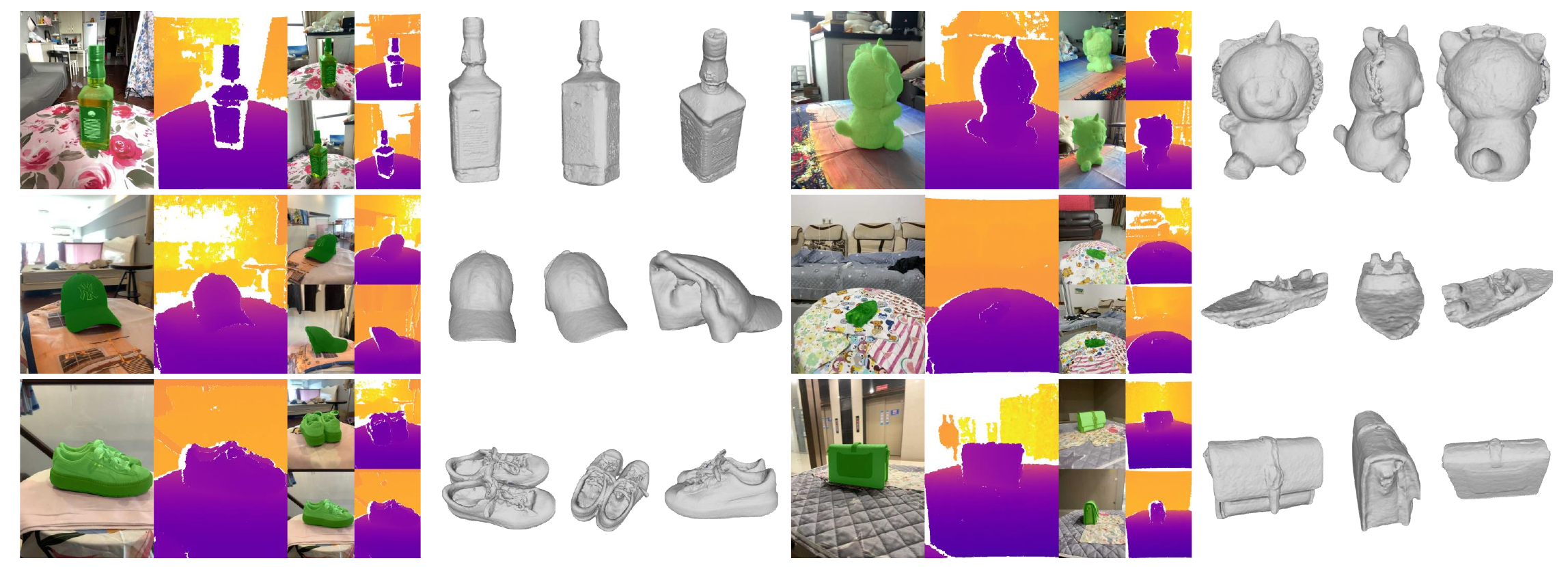}
   
   \end{tabular}
   
   }
   \vspace{-10px}
   \captionof{figure}{{\bf Visualization of RGBD reconstruction surface} {from Neusfacto~\cite{Yu2022SDFStudio}. Original RGBD image samples are listed on the left and multi-view reconstructed surface is on the right for each example.}}
    \label{fig:recon_vis}
\end{table*}

\vspace{-12px}
\paragraph{Depth Supervised NVS}
We also study the influences that depth supervision brings. We choose Instant-NGP~\cite{muller2022instant} in single-scene NVS methods and both Pixel-NeRF~\cite{yu2021pixelnerf} and MVSNeRF~\cite{chen2021mvsnerf} in cross-scene NVS methods. Our experiment results in Tab.~\ref{tab:depth_sup_nvs} prove that depth supervision is beneficial for these methods to learn better representations. In our experiment setting, we add L1 depth loss to every algorithm and choose the best-performance depth loss weight for them. Compared with conducting NVS tasks without depths, the performances of both Instant-NGP and Pixel-NeRF get boosted when training with depth loss. As for experiments of MVSNeRF, since we've already added depth information in the original training as the guidance in building rays, it turns out that the improvements when training with extra added depth loss are limited. In a nutshell, with depth priors, both single-scene NVS and cross-scene NVS methods learn better generalization capabilities, boosting NVS more accurate and generalizable.

\subsection{Camera Pose Estimation}
In this section, we benchmark two data-driven methods RelPose~\cite{zhang2022relpose} and RelPose++~\cite{lin2023relpose++} for inference of the relative camera poses from multi-view images in a sparse setting. Leveraging the given annotations of camera poses and large-scale category-level video in our dataset, we aim to learn generalizable viewpoint inference capability from training-seen categories to unseen ones. Since \datasetname~dataset has a full and dense 360-degree camera trajectory, we can provide both a large-scale database and various view settings to assist training. 
In our experiments, we divided totally 46 categories into training and testing categories. We also hold some videos in training categories for evaluation. We adopt evaluation settings described in \cite{zhang2022relpose, lin2023relpose++} and report results in Tab.~\ref{tab:relpose} and Tab.~\ref{tab:relposepp}. We observe that these two methods can generalize well to other scenarios both in known categories and unseen categories since the relative rotation estimation errors are in a reasonable range (also see Figure~\ref{fig:relpose} for visualization). However, the error of translation prediction is comparatively large in RelPose++, which still poses challenges in this field.
\sumbullet{To sum up, \datasetname~dataset can serve as large-scale training sources for generalizable camera pose estimation algorithms to achieve remarkable results.}

\subsection{RGBD Object Surface Reconstruction}
In our experiment setting of object surface reconstruction, algorithms need to utilize RGBD image sequence and central object masks to reconstruct the surface mesh of the central object. Reconstruction without depths is also evaluated for comparison. For evaluation of reconstruction quality, we calculate the Chamfer Distance between the reconstruction mesh and aggregated object point cloud which is derived from object-masked depth images.  Ten single object scenes are selected in each category for evaluations of Instant-NGP~\cite{muller2022instant} and Neusfacto~\cite{Yu2022SDFStudio}.  From the results in Tab.~\ref{tab:rgbd_recon}, we observe that reconstruction is better with depth priors. Additionally, the performance of Neusfacto with RGBD is superior to Instant-NGP, which shows that depths help the sdf-based method Neusfacto learn the correct object boundary and boost the performance more compared with Instant-NGP. The average deviation is high due to the varied reconstruction qualities across different categories in the dataset. Visualization of Neusfacto RGBD reconstructions are shown in Figure~\ref{fig:recon_vis}. \sumbullet{In brief, our dataset provides an RGBD object reconstruction evaluation track, boosting the development of more mature algorithms in this field.}

\begin{table}
   \centering
   \resizebox{\linewidth}{!}{
   
   \begin{tabular}{cccc}
   \includegraphics[width=0.4\linewidth]{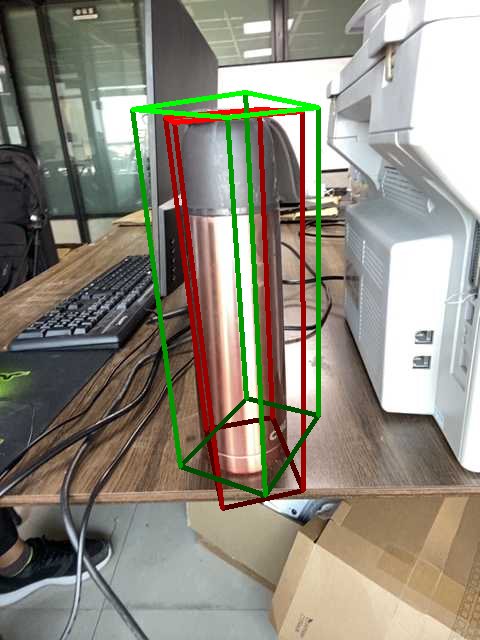} &
   \includegraphics[width=0.4\linewidth]{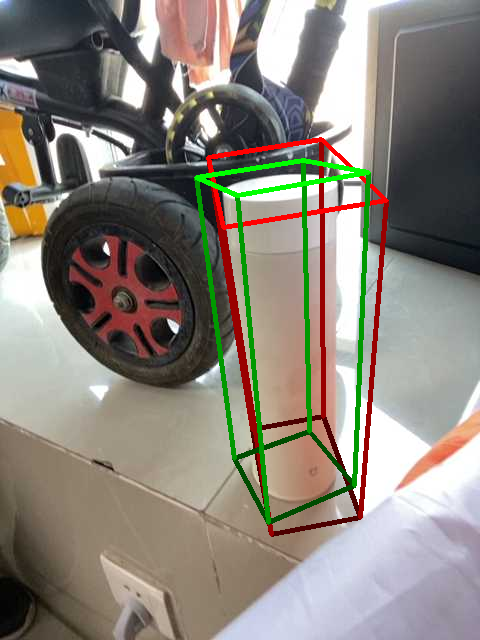} &
   \includegraphics[width=0.4\linewidth]{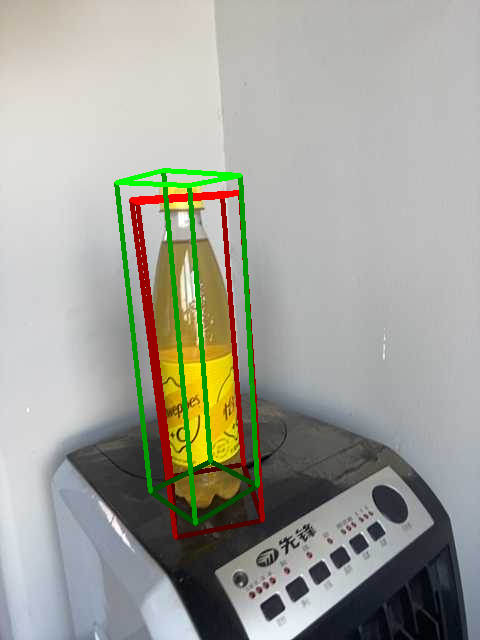} &
   \includegraphics[width=0.4\linewidth]{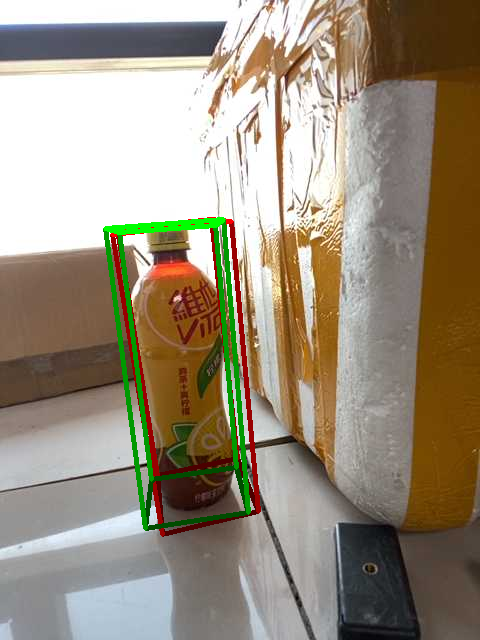} \\
\includegraphics[width=0.4\linewidth]{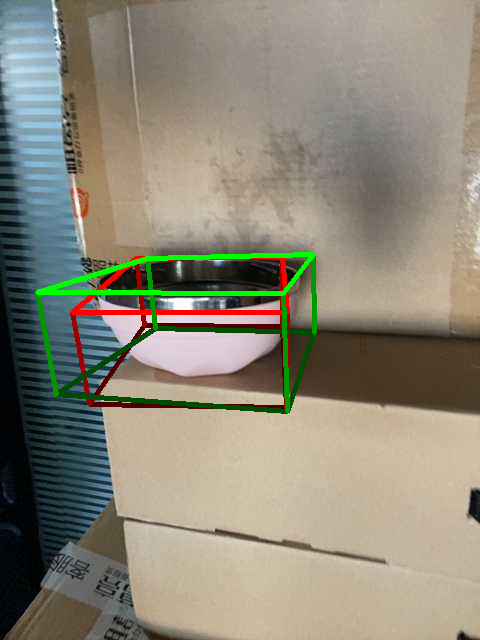} &
   \includegraphics[width=0.4\linewidth]{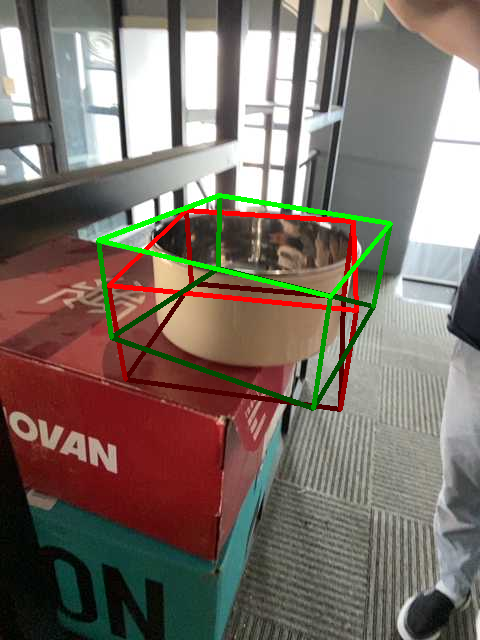} &
   \includegraphics[width=0.4\linewidth]{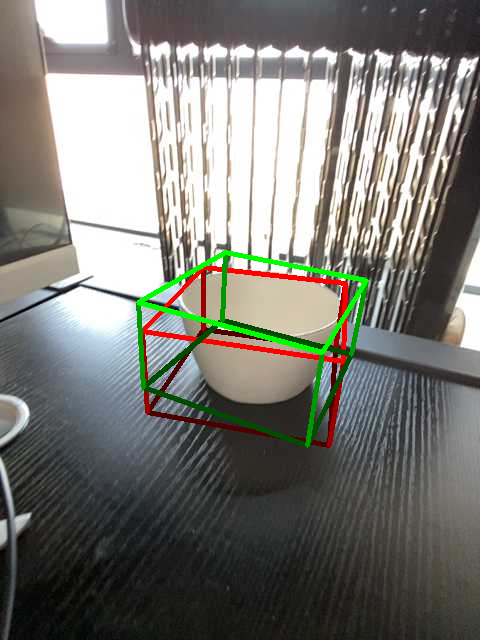} &
   \includegraphics[width=0.4\linewidth]{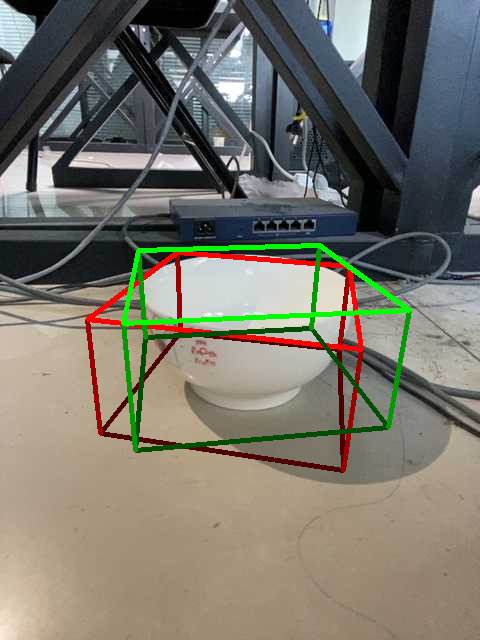} \\
\includegraphics[width=0.4\linewidth]{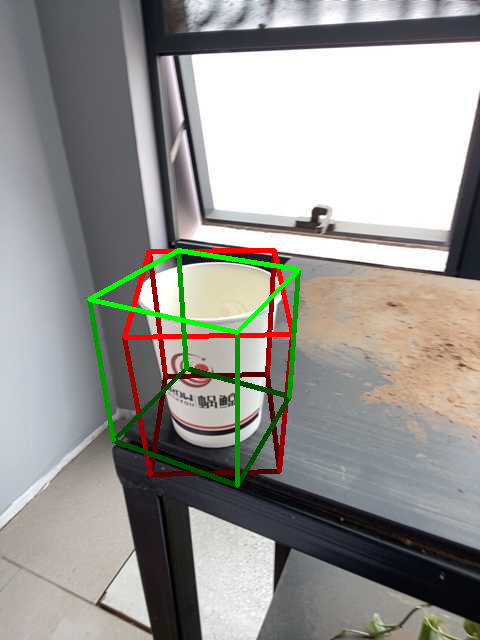} &
   \includegraphics[width=0.4\linewidth]{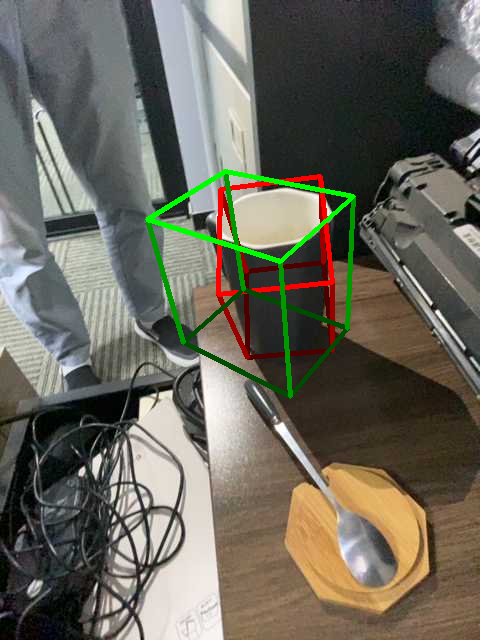} &
   \includegraphics[width=0.4\linewidth]{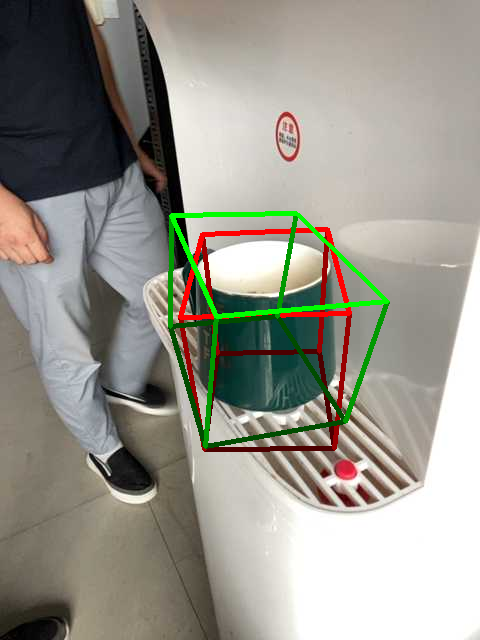} &
   \includegraphics[width=0.4\linewidth]{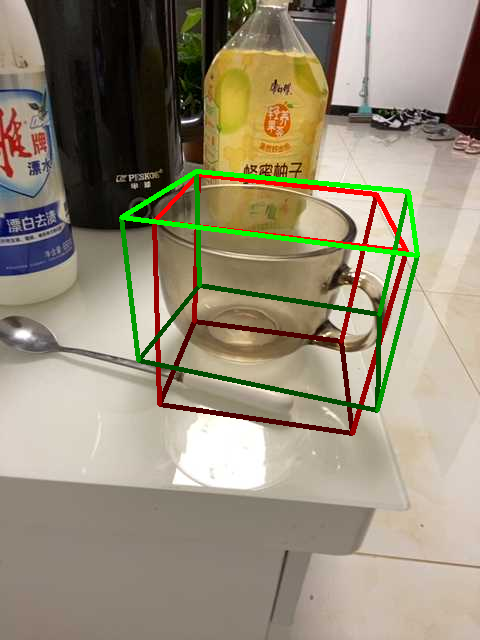} \\
   
   \end{tabular}
   }
   \vspace{-5px}
   \captionof{figure}{{\bf Object 6D pose estimation visualization.} {We visualize the predicted category-level 6D pose on 
 three categories in Wild6D~\cite{fu2022categorylevel} test set (bottle, bowl, and mug) using models that \textbf{only} perform self-supervised training on the corresponding category of \datasetname~Dataset. The ground truth bounding boxes are colored in green, and the predicted bounding boxes are in red.}}
    \label{fig:6dpose}
\end{table}
\begin{table}[t]

    \resizebox{0.48\textwidth}{!}{
    \centering
    \begin{tabular}{c|c|c|cccc}
    \toprule
        \multirow{2}{*}{Eval. Type} & \multirow{2}{*}{Categories}  & \multirow{2}{*}{Metrics} & \multicolumn{4}{c}{\#Frames} \\
        & & & 3 & 5 & 10 & 20 \\
        \midrule
        \multirow{4}{*}{MST} & \multirow{2}{*}{seen} & $<$15 deg. & 57.4 & 55.1 & 51.4 & 47.4 \\
        & &  $<$30 deg. & 82.1 & 79.8 & 77.2 & 74.0 \\
        & \multirow{2}{*}{unseen} &  $<$15 deg. & 38.4 & 37.7 & 36.6 & 35.2 \\
        & &  $<$30 deg. & 62.5 & 61.8 & 60.4 & 59.0 \\
        \midrule
        \multirow{4}{*}{Coord.Asc.} & \multirow{2}{*}{seen} & $<$15 deg. & 69.3 & 69.3 & 69.8 & 69.3\\
        & &  $<$30 deg. & 85.3 & 85.3 & 85.4 & 85.3 \\
        & \multirow{2}{*}{unseen} &  $<$15 deg. & 46.0 & 46.2 & 46.9 & 46.5\\
        & &  $<$30 deg. & 66.4 & 67.0 & 67.2 & 67.1\\
        \midrule
        \multirow{4}{*}{Sequential} & \multirow{2}{*}{seen} & $<$15 deg. &  51.9 &  45.1 &  36.0 & 26.9 \\
        & &  $<$30 deg. & 78.3 & 72.5 & 61.9 & 49.3 \\
        & \multirow{2}{*}{unseen} &  $<$15 deg. & 34.9 & 31.0 & 25.1 & 18.3 \\
        & &  $<$30 deg. & 59.1 & 54.4 & 46.4 & 37.1 \\
    \bottomrule
    \end{tabular}
    }
\caption{\textbf{RelPose~\cite{zhang2022relpose} camera evaluation results.} {We follow three evaluation types (MST, Coord.Asc., Sequential) propsed in \cite{zhang2022relpose} and report the average percent of rotation prediction errors in degrees both in training-seen categories and unseen ones. }}
    \label{tab:relpose}
\end{table}
\begin{table}[t]
    \resizebox{0.48\textwidth}{!}{
    \centering
    \begin{tabular}{c|c|c|cccc}
    \toprule
        \multirow{2}{*}{Eval. Type} & \multirow{2}{*}{Categories}  & \multirow{2}{*}{Metrics} & \multicolumn{4}{c}{\#Frames} \\
        & & & 2 & 3 & 5 & 8 \\
        \midrule
        \multirow{4}{*}{Pairwise} & \multirow{2}{*}{seen} & $<$15 deg. & 69.6 & 68.3 & 67.3 & 66.6 \\
        & &  $<$30 deg. & 86.5 & 87.4 & 87.2 & 86.8 \\
        & \multirow{2}{*}{unseen} &  $<$15 deg. & 53.4 & 52.5 & 52.5 & 52.3 \\
        & &  $<$30 deg. & 74.1 & 74.5 & 75.3 & 75.4 \\
        \midrule
        \multirow{4}{*}{Coord.Asc.} & \multirow{2}{*}{seen} & $<$15 deg. & 70.4 & 71.5 & 71.9 & 71.6 \\
        & &  $<$30 deg. & 86.7 & 87.9 & 88.5 & 88.7 \\
        & \multirow{2}{*}{unseen} &  $<$15 deg. & 52.9 & 55.4 & 
54.9 & 54.8 \\
        & &  $<$30 deg. & 73.9 & 75.7 & 76.3 & 76.8 \\
        \midrule
        \multirow{4}{*}{Cam.Center} & \multirow{2}{*}{seen} & $<$0.2 SS & 100.0 & 29.8  & 12.6 & 5.9 \\
         &  & $<$0.3 SS & 100.0 & 43.8 & 23.6 & 13.3 \\
         & \multirow{2}{*}{unseen} & $<$0.2 SS & 100.0  & 30.5 & 12.4 & 5.6 \\
         &  & $<$0.3 SS & 100.0 & 44.3 & 23.2 & 12.5 \\
        \midrule
        \multirow{4}{*}{Cam.Trans.} & \multirow{2}{*}{seen} & $<$0.2 SS & 22.3 & 11.9 & 4.8 & 2.2 \\
         &  & $<$0.3 SS & 30.1 & 20.4 & 10.8 & 6.2 \\
         & \multirow{2}{*}{unseen} & $<$0.2 SS & 21.8 & 12.0 & 5.7 & 2.7 \\
         &  & $<$0.3 SS & 29.8 & 20.9 & 12.0 & 6.4 \\
    \bottomrule
    \end{tabular}
    }
    \caption{\textbf{RelPose++~\cite{lin2023relpose++} camera evaluation results.} {We follow four evaluation types (Pairwise, Coord.Asc., Cam.Center, Cam.Trans) proposed in \cite{lin2023relpose++} and report the average percent of rotation prediction errors in degrees both in training-seen categories and unseen ones. Notes: SS means scene scale defined in \cite{lin2023relpose++}.}}
    \label{tab:relposepp}

\end{table}

\begin{table}[t]
    \centering
    \resizebox{0.45\textwidth}{!}{
    \begin{tabular}{c|c|c|c}
    \toprule
        \multicolumn{2}{c}{RGB} & \multicolumn{2}{|c}{RGBD} \\ 
        \midrule
        Instant-NGP~\cite{muller2022instant} & Neusfacto~\cite{Yu2022SDFStudio} &
        Instant-NGP~\cite{muller2022instant} & Neusfacto~\cite{Yu2022SDFStudio} \\
        \midrule
        45.91/64.01 & 88.92/89.94 & 28.46/29.28 & \textbf{25.83}/34.07 \\
    \bottomrule
    \end{tabular}
    }
    \caption{\textbf{RGBD object surface reconstruction results.} {Average of chamfer distance with standard deviation across selected categories are reported (Average/SD).}}
    \label{tab:rgbd_recon}

\end{table}
\begin{table}[t]

    \resizebox{0.48\textwidth}{!}{
    \centering
    \begin{tabular}{l|c|cc|cccc}
    \toprule
        \multirow{2}{*}{Category} & \multirow{2}{*}{Datasets} &  \multirow{2}{*}{IOU$_{0.25}$} & \multirow{2}{*}{IOU$_{0.5}$} & 5 deg. & 5 deg. & 10 deg. & 10 deg. \\
         &  &   &  & 2cm & 5cm & 2cm & 5cm \\
        \midrule
        \multirow{3}{*}{Bottle} & Wild6D & 93.2 & 85.2 & 71.3 & \textbf{79.4} & 79.8 & 90.9 \\
         & ROW & \textbf{93.3} & 70.9 & 34.1 & 48.8 & 47.9 & 78.8 \\
         & Wild6D+ROW & \textbf{93.3} & \textbf{85.8} & \textbf{71.9} & 78.6 & \textbf{81.7} & \textbf{91.7} \\
        \midrule
         \multirow{3}{*}{Bowl} & Wild6D & 98.3 & 90.4 & \textbf{66.1} & \textbf{70.0} & 86.8 & \textbf{94.6} \\
         & ROW & 98.3 & \textbf{91.8} & 33.8 & 35.7 & 86.3 & 93.5 \\
         & Wild6D+ROW & \textbf{98.4} & \textbf{91.8} & 40.3 & 42.0 & \textbf{87.5} & 93.7 \\
        \midrule
         \multirow{3}{*}{Mug} & Wild6D & 89.0 & 59.2 & 0.0 & 0.0 & 0.0 & \textbf{0.1} \\
         & ROW & 89.1 & \textbf{61.9} & 0.0 & 0.0 & 0.0 & \textbf{0.1} \\
         & Wild6D+ROW & \textbf{89.3} & 50.2 & 0.0 & 0.0 & 0.0 & 0.0 \\
    \bottomrule
    \end{tabular}    
    }
    \caption{\textbf{Self-Supervised 6D Pose Estimation results.} {The evaluation results on Wild6D test dataset under three different settings in bottle, bowl and mug categories.}}
    \label{tab:6dpose}

\end{table}
\subsection{RGBD 6D Object Pose Estimation}
We explore our dataset in self-supervised training for 6D pose estimation. We adopt the algorithm proposed in~\cite{zhang2022self}, which leverages category shape prior and learns by matching the correspondence between images and shapes. In our experiment, we evaluate the trained model on Wild6D~\cite{fu2022categorylevel} test set. Three different settings concerning training sets are adopted: 1) Wild6D training set; 2) \datasetname~dataset; 3) Wild6D training set + \datasetname~dataset. Common categories in these two datasets are selected for self-supervised training and evaluation. Results in Tab.~\ref{tab:6dpose} show that in the case of an out-of-distribution setting, where we trained only on our dataset and evaluated on a different dataset, some of the metrics are decreased in our experiments. This is mainly due to the different distribution of camera rotations in these two datasets (visualized in Fig.~\ref{fig:distvis}), where Wild6D doesn't cover full 360 degrees and \datasetname~dataset covers a larger pitch angle range in object 6D poses. However, we still witness some improvements in evaluations. Training with \datasetname~dataset benefits in IOU evaluations and joint-dataset training improves rotation+translation evaluation in particular categories. Visualization of 6D pose estimation in Wild6D test set using models that only train in our dataset can be found in Figure~\ref{fig:6dpose}, To summarize, \sumbullet{our dataset provides large-scale category-level RGBD images sequences, serving as ample unsupervised training data, which has the potential to boost more accurate 6D pose estimation in the future.}

\section{Conclusion}

The object-centric datasets in the computer vision community have mostly focused on RGB videos, while practical applications often involve depth as inputs or for better annotations. We collect the largest object-centric RGB-D video dataset \datasetname~, where all videos are captured in cluttered scenes. It is composed of category-level RGB-D object videos taken using iPhones around the objects in 360 degrees, which contains around 8500 recorded objects and nearly 20000 RGB-D videos across 46 common object categories with three setups covering most scenarios. The dataset is well annotated with object masks, real-world scale camera poses, and reconstructed aggregated point clouds from RGBD videos. We set up four evaluation tracks with \datasetname, showing that the large-scale capture of RGB-D objects provides a large potential to advance 3D object learning. The current dataset does not come with annotations of the object 6D pose, which requires further crowd-sourcing effort. It will be one of our future efforts to collect this annotation for supervised training methods as well as evaluation. We are committed to releasing our dataset and evaluation code.

\paragraph{Acknowledgment} This project was supported, in part, by the Amazon Research Award, the Qualcomm Innovation Fellowship, the Intel Rising Star Faculty Award, and the CISCO Faculty Award.

{
    \small
    \bibliographystyle{ieeenat_fullname}
    \bibliography{main}
}


\end{document}


\clearpage
\setcounter{page}{1}
\maketitlesupplementary

\appendix
\section{Additional Details of WildRGB-D Dataset}
\paragraph{Video types distribution}

Our dataset comprises objects paired with their respective videos, with each object being associated with three recorded videos. One of these videos is labeled as the \emph{Single object video}, while the classification of the other two videos is contingent on the nature of the object. In our data collection strategy, we selectively opt for certain object categories to capture  \emph{Hand-object video}, excluding categories where objects, such as toy trains or buses, lack meaningful interactions with hands. Within our chosen categories, no more than 10 percent of objects feature a single clip of  \emph{Hand-object video}, and the remaining videos consistently fall into the category of \emph{Multi-object video}.

\paragraph{Video recording details}
To ensure comprehensive 360-degree recording with minimal camera shaking and motion blurs in our videos, we implement a turntable hidden beneath the table, out of the camera's view. This turntable features an adjustable arm that can vary the pitch angle and distance. At the end of the arm, there's a mount for an iPhone, facilitating recording. During filming, we can rotate the arm, ensuring a smooth and uniform rotation for stable footage.
To enhance dataset diversity, we deliberately choose various radii and pitch angles during recording. Additionally, for the background of objects, we deliberately select diverse scenarios to further enrich the variability of our dataset. The average video length is over 600 frames.

\paragraph{Dataset mask annotation details.} With Grounding-DINO~\cite{liu2023grounding} to get prompts (bounding box) for SAM~\cite{kirillov2023segment}, every generated mask will be examined again by the annotators to ensure mask quality. If the mask is wrong, we will label the mask manually by explicit clicks in the image, which serve as click prompts for SAM, and generate the final correct mask. This generates almost the same quality of results as manually labeling all masks but is much more efficient.

\paragraph{Video quality examination}
We systematically exclude videos that fail to meet our predefined criteria. Specifically, any videos lacking a complete 360-degree recording are eliminated from consideration. For the quality of captured depths, errors of Apple's TrueDepth Camera only reach up to 5\% of the target distance, so the quality of depths is quite good in our collected RGB-D videos. After applying Simultaneous Localization and Mapping (SLAM) processing to determine camera poses, we discard videos that exhibit an unreasonable camera trajectory. This evaluation is conducted by meticulous examination of point cloud reconstructions and visualized camera trajectories. Through this rigorous quality control process, we guarantee that our retained videos exhibit a consistently high level of quality.
\begin{table}[]
\vspace{-0.1in}
    \resizebox{0.48\textwidth}{!}{
    \centering
    \begin{tabular}{l|cc|cc}
    \toprule
        \multirow{2}{*}{COLMAP Result} & \multicolumn{2}{c|}{RGB} & \multicolumn{2}{c}{RGB + Depth Sup} \\
        & COLMAP & RGBD SLAM & COLMAP & RGBD SLAM\\
        \hline
        79/138 (\textbf{57.2\%}) Fail & \textcolor{red}{N/A} & 29.85 & \textcolor{red}{N/A} & 30.86 \\
        59/138 (42.8\%) Success & 31.10 &  31.33 & 31.75 & 31.38 \\
    \bottomrule
    \end{tabular}    
    }
    \caption{\textbf{The results of view synthesis} using different ways to estimate camera poses. We report the PSNR which is evaluated with NeuS2~\cite{wang2023neus2}.}
    \label{tab:neus2}
\end{table}
\paragraph{Evaluation of pose quality from COLMAP and SLAM} To evaluate the accuracy of camera poses estimated from COLMAP~\cite{schonberger2016structure} and RGBD SLAM~\cite{durrant2006simultaneous,schops2019bad}, we run NeuS2~\cite{wang2023neus2} with camera poses estimated from them, and report the PSNR of view synthesis in  Tab.~\ref{tab:neus2}. Column 2 and 3 show results using only RGB supervision; Column 3 and 4 use both RGB and depth as supervision. We randomly choose three single object scenes from every category in our dataset, resulting in 138 scenes. We find COLMAP fails in {57.2\%} of the scenes on camera pose estimation, which RGBD SLAM can succeed in all cases. This shows \emph{the significance of how much depth can help} in camera pose estimation. In the cases where COLMAP fails  (1st row), adding depth supervision can boost view synthesis by a large margin with RGBD SLAM cameras. We report N/A with COLMAP cameras as the estimations fail. In 2nd row where COLMAP works, all ablations lead to similar results. This suggests depth is very useful in challenging scenes that cover a significant number of data.

\paragraph{Personal data and human subject}
During our dataset collection process, we enlisted the assistance of hired workers to record videos on our behalf. In a small fraction of these videos, the workers inadvertently appear, featuring their hands or partial bodies. Importantly, we have secured explicit consent from these individuals to include these specific portions of the videos in our dataset. This ensures that the inclusion of worker-related content is both intentional and authorized, maintaining transparency and adherence to ethical standards in our dataset compilation.


\section{Additional Experiment Details}
\subsection{Novel view synthesis}

\paragraph{Dataset splits}
In the context of Single-Scene Novel View Synthesis (NVS), where NeRF-based methods~\cite{mildenhall2020nerf, barron2022mip, muller2022instant} are evaluated, we employ a randomized approach wherein we select ten scenes at random from each category. Subsequently, we uniformly subsample each video to a fixed length of 100 frames and further extract 20 percent of these frames for validation purposes. For Cross-Scene NVS experiments, we designate the same ten scenes chosen for the Single-Scene NVS as the test scenes within each category. The remaining scenes in each category are then utilized for training. During evaluations, we exclusively test on the 20 percent of images sampled from these test video clips, mirroring the methodology employed in Single-Scene NVS experiments. The remaining images serve as source views for synthesizing renderings in the context of Generalizable Neural Radiance Fields (NeRFs)~\cite{yu2021pixelnerf, chen2021mvsnerf, wang2021ibrnet}. Notably, for every method, we consistently employ three source views for the synthesis of renderings, which are chosen in a deterministic way. Difficulty level division in Cross-Scene NVS experiments is caused by various object shapes, backgrounds, and the number of training videos in each category.
\paragraph{Training details} For every NeRF-based method, we largely follow its original training process. NeRF~\cite{mildenhall2020nerf}, MipNeRF 360~\cite{barron2022mip} and Instant-NGP~\cite{muller2022instant} are all trained with 30k iterations using their default hyper-parameters. Pixel-NeRF~\cite{yu2021pixelnerf}, MVSNeRF~\cite{chen2021mvsnerf} and IBRNet~\cite{wang2021ibrnet} are trained using default settings, but with different iterations and epochs. Pixel-NeRF~\cite{yu2021pixelnerf} is trained in 200 epochs, MVSNeRF~\cite{chen2021mvsnerf} and IBRNet~\cite{wang2021ibrnet} are trained with 100k iterations. We ensure enough training time for every method to be correctly evaluated.
\subsection{Camera pose estimation}
\paragraph{Dataset splits}
Our dataset, consisting of 46 categories, is systematically divided into 27 training categories and 19 test categories. During the training phase, we exclusively utilize 70 percent of the videos from the training categories, reserving the remaining 30 percent for validation. The metrics derived from this validation process are reported under the designation of ``seen" categories. Evaluation results on the test categories are distinctly reported as ``unseen" categories. This partitioning strategy ensures a robust assessment of model performance on both familiar and novel categories, contributing to a comprehensive evaluation framework. For every video, we uniformly subsample each video to a fixed length of 100 frames as well.
\paragraph{Training details}
We follow the training procedure and hyper-parameters in RelPose~\cite{zhang2022relpose} and RelPose++~\cite{lin2023relpose++}, with different iterations: RelPose~\cite{zhang2022relpose} is trained with 100k iterations and RelPose++~\cite{lin2023relpose++} with 400k iterations.
\subsection{Object surface reconstruction}
\paragraph{Implementation details}
For Instant-NGP~\cite{muller2022instant}, we adopt its re-implementation in \cite{ngpimpl}. We apply an additional mask loss. To be more specific, for every casting ray $\mathbf{r}(t) = \mathbf{o} + t\mathbf{d}$ from the camera center $\mathbf{o}$ through the pixel center in direction $\mathbf{d}$,  Instant-NGP samples a set of 3D points $\{\mathbf{x}_i\}$ along the ray. After querying respective density $\{\sigma_i\}$ of the points $\{\mathbf{x}_i\}$, we calculate the opacity of the ray by $\sum_i w_i$. We want to make it aligned with the mask $m_r$ of that ray by the mask loss $\mathcal{L}_\text{mask} =\sum_\mathbf{r} \| \sum_i w_i - \mathbf{m}_r \|^{2}_2$. We additionally add depth loss to Instant-NGP in RGB-D surface reconstruction. For Neusfacto~\cite{Yu2022SDFStudio}, we follow their implementations in SDFStudio, applying their proposed mask loss and sensor depth loss. 
\paragraph{Training details}
We train Instant-NGP~\cite{muller2022instant} with 30k iterations. For RGB surface reconstruction in Neusfacto~\cite{Yu2022SDFStudio}, we train with a longer 60k iterations. RGB-D surface reconstruction in Neusfacto~\cite{Yu2022SDFStudio}, we train with only 10k iterations because of its fast convergence speed.

\subsection{Object 6D pose estimation}
\paragraph{Training details} We choose three common categories of our dataset and Wild6D~\cite{fu2022categorylevel} for category-level 6D pose estimation. We don't subsample our video here. We train every model following \cite{zhang2022self} for 20k iterations with default hyperparameters. 
{
    \small
    \bibliographystyle{ieeenat_fullname}
    \bibliography{main}
}
